\documentclass[10pt,twocolumn,letterpaper]{article}

\usepackage{cvpr}
\usepackage{times}
\usepackage{epsfig}
\usepackage{graphicx}
\usepackage{amsmath}
\usepackage{amssymb}

\def\x{{\mathbf x}}

\usepackage[pagebackref=true,breaklinks=true,letterpaper=true,colorlinks,bookmarks=false]{hyperref}

\cvprfinalcopy 

\def\cvprPaperID{0009} 

\ifcvprfinal\fi
\begin{document}

\title{Recurrent Memory Addressing for describing videos}

\author{Arnav Kumar Jain\thanks{denotes equal contribution}\hspace{5mm} Abhinav Agarwalla$^{*}$ \hspace{5mm} Kumar Krishna Agrawal$^{*}$ \hspace{5mm}  Pabitra Mitra\\
Indian Institute of Technology Kharagpur\\
{\tt\small \{arnavkj95, abhinavagarawalla, kumarkrishna, pabitra\}@iitkgp.ac.in}
}
\maketitle

\begin{abstract}
In this paper, we introduce Key-Value Memory Networks to a multimodal setting and a novel key-addressing mechanism to deal with sequence-to-sequence models. The proposed model naturally decomposes the problem of video captioning into vision and language segments, dealing with them as key-value pairs. More specifically, we learn a semantic embedding (v) corresponding to each frame (k) in the video, thereby creating (k, v) memory slots. We propose to find the next step attention weights conditioned on the previous attention distributions for the key-value memory slots in the memory addressing schema. Exploiting this flexibility of the framework, we additionally capture spatial dependencies while mapping from the visual to semantic embedding. Experiments done on the Youtube2Text dataset demonstrate usefulness of recurrent key-addressing, while achieving competitive scores on BLEU@4, METEOR metrics against state-of-the-art models.
\end{abstract}

\section{Introduction} \label{intro}
Generating natural language descriptions for images and videos is a long-standing problem, in the intersection of computer vision and natural language processing. Solving the problem requires developing powerful models capable of extracting visual information about various objects in an image, while deriving semantic relationships between them in natural language. For video captioning, the models are additionally required to find compact representations of the video to capture the temporal dynamics across image frames.

\begin{figure}
\centering
\includegraphics[scale=0.45]{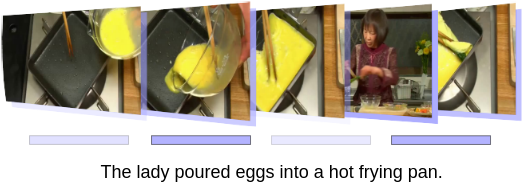}
\caption{Our model employs a temporal attention mechanism on the visual features to identify key frames in the video. These are mapped to semantic features in the language domain for better context to the language model, which then generates the output sequence. Previously attended frames and generated words identify the key frames for generating the next word.}
\label{fig:intro}
\end{figure}

The recent advances in training deep neural architectures have significantly improved in the state-of-the-art across computer vision and natural language understanding. With impressive results in object detection and scene understanding, Convolution Neural Networks (CNNs) \cite{lecun1998gradient} have become the staple for extracting feature representations from images. Recurrent Neural Networks (RNNs) with Long Short Term Memory (LSTM) \cite{hochreiter1997long} units or Gated Recurrent Units (GRUs)\cite{chung2014empirical}, have similarly emerged as generative models of choice for dealing with sequences in domains ranging from language modeling, machine translation to speech recognition. Advancements in these fundamental problems make tackling challenging problems, like captioning \cite{johnson2015densecap,DBLP:journals/corr/XuBKCCSZB15}, dialogue \cite{serban2016building} and visual question answering \cite{antol2015vqa} more viable.

Despite the fundamental complexities of these problems, there has been an increasing interest in solving them. A common underlying approach in these proposed models is the notion of "attention mechanisms", which refers to selectively focusing on segments of sequences \cite{bahdanau2014neural,yao2015describing} or images \cite{stollenga2014deep} to generate corresponding outputs. Such attention based approaches are specially attractive for captioning problems, since they allow the network to focus on patches of the image conditioned on the previously generated tokens \cite{DBLP:journals/corr/XuBKCCSZB15, johnson2015densecap}, often referred to as spatial attention. 

Models with spatial attention, however cannot be readily used for video description. For instance, in the Youtube2Text dataset, a video clip stretches around 10 seconds, or around 150 frames. Applying attention on patches in these individual frames provides the network with local spatial context. This however, does not take ordering of the frame sequence or events ranging across frames, into consideration. To incorporate this \textit{temporal attention} into the model, \cite{yao2015describing,yu2015video,pan2015hierarchical} extend this \textit{soft alignment} to video captioning. Most of these approaches, treat the problem of video captioning in the sequence-to-sequence paradigm \cite{sutskever2014sequence} with attentive encoders and decoders. This requires finding a compact representation of the video, which is passed as context to the RNN decoder. 

However, we identify two primary issues with these approaches. First, applying attention sequentially provides the model with local context at the generative decoder \cite{yang2016encode}. As a result the decoder would be unable to deal with long-term dependencies while captioning videos of longer duration. Secondly, these models jointly learn the multimodal embedding in a visual-semantic space \cite{pan2015jointly,yu2015video} at the RNN decoder. With the annotated sentences being the only supervisory signal, learning a mapping from a sequence of images to a sequence of words is difficult. This is specially true for dealing with video sequences, as the underlying probability distribution is distinctively multimodal.  While \cite{pan2015jointly} tries to address this issue with an auxiliary loss, the model suffers from the first drawback.

To address the aforementioned issues, we introduce a model which generalizes Key-Value Memory Networks \cite{miller2016key} to a multimodal setting for video captioning. At the same time, the framework provides an effective way to deal with the complex transformation from the visual to language domain. Using a pre-trained model to explicitly transform individual frames (\textit{keys}) to semantic embedding (\textit{values}), we construct memory slots with each slot being a tuple (\textit{key, value}). This allows us to provide a weighted pooling of textual features as context to the decoder RNN, which is closer to the language model. The proposed model naturally tackles the problem of maintaining long-term temporal dependencies in videos, by explicitly providing all image frames for selection at each time step. We also propose a novel key-addressing scheme (see Section \ref{keyadd}), allowing us to find the new relevance scores conditioned on the previous attention distribution. It keeps track of previous attention distribution and provides a global context to the decoder. This allows us to exploit the temporal dependencies in the recurrent key-addressing and in the language decoder.

In summary, our key contributions are following:
\begin{itemize}
\item We generalize Key-Value Memory Networks (KV-MemNN) in a multimodal setting to generate natural descriptions for videos, and more generally deal with sequence-to-sequence models (Section \ref{keyval}).
\item We propose a novel key-addressing schema to find the attention weights for key-value memory slots conditioned on the previous attention distribution (Section \ref{keyadd}).
\item The proposed model is evaluated on the YouTube dataset \cite{chen2011collecting}, where we outperform strong baselines while reporting competitive results against state-of-art models (Section \ref{results}).
\end{itemize}


\section{Related Work}
Following the success of end-to-end neural architectures and attention mechanisms, there is a growing body of literature for captioning tasks, in images and more recently videos. To deal with the multimodal nature of the problem, classical approaches relied on manually engineered templates \cite{kulkarni2013babytalk,das2013thousand}. And while some recent approaches in this direction show promise \cite{fang2015captions}, but the models lack generalization to deal with complex scenes, videos. 

As an alternative approach, \cite{frome2013devise, kiros2014unifying} suggest learning a joint visual-semantic embedding, effectively a mapping from the visual to language space. The motivation of our work is strongly aligned with \cite{rohrbach2013translating}, who generate semantic representations for images using CRF models, as context for the language decoder. However, our approach significantly differs in the essence that we capture spatio-temporal dynamics in videos while generating the text description.  

Building on this, and Encoder-Decoder \cite{bahdanau2014neural, cho-al-emnlp14} models for machine translation , \cite{vinyals2015show, venugopalan:naacl15} develop models which compute average fixed-length representations (for images, videos respectively) from image features. These \textit{context} vectors are provided at each time step to the decoder language model, for generating descriptions. The visual representations for the images are usually transferred from pretrained convolution networks \cite{simonyan2014very, szegedy2015going}.

A major drawback of the above approach is induced by mean pooling, where context features across image frames are collapsed. For one, this looses the temporal structure across frames by treating them as "bag-of-images" model. Addressing this, \cite{sutskever2014sequence} propose Sequence-to-Sequence models for accounting for the temporal structure, and \cite{venugopalan15iccv} extend it to a video-captioning setting. However, passing a fixed vector as context at each time step, creates a bottleneck for the flow of gradients using Backpropagation Through Time (BPTT) \cite{werbos1990backpropagation} at the encoder.

\begin{figure*}
\centering
\includegraphics[scale=0.55]{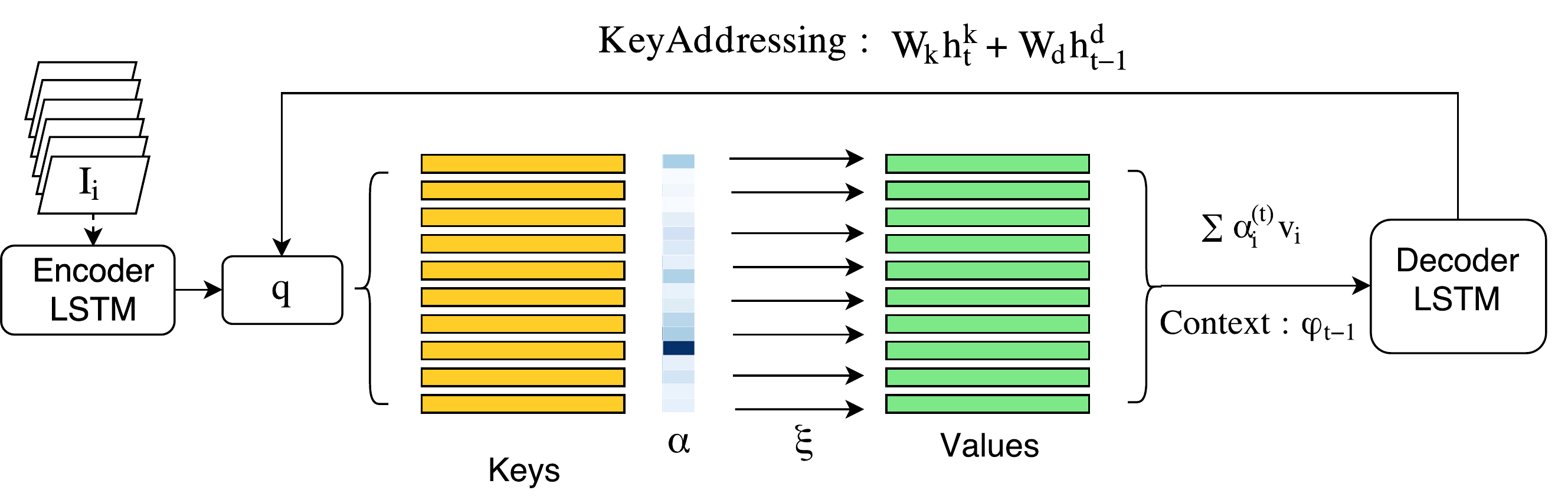}
\caption{The video is considered as a sequence of image frames \{$I_1, ..., I_n$\}. The memory is filled with key-value pairs ($k_i, v_i$) that capture the relationships between visual features and textual descriptions. The $\alpha^t_i$ corresponds to the attention weights associated with the memory slots ($h_{t-1}, h_t, ... $) being the hidden states of the decoder RNN. The memory is then queried over and over again to produce a weighted sum of the values to be decoded using a standard LSTM RNN decoder into a word in the description.}
\label{fig:vidmem}
\end{figure*}

The notion of \textit{visual attention} has a rich literature in Psychology and Neuroscience, and has recently found application in computer vision \cite{mnih2014recurrent} and machine translation \cite{bahdanau2014neural}. Allowing the network to selectively focus on the patches of images or segments of the input sequences, representative works \cite{DBLP:journals/corr/XuBKCCSZB15, yao2015describing,johnson2015densecap, ballas2015delving,yu2015video,pan2015hierarchical,pan2015jointly} have significantly pushed the state-of-the-art in their domain. The issues of fixed length representation and gradient bottleneck are largely addressed by selectively conditioning the decoder outputs on encoder states:
\begin{equation}
p(y_i | y_{i-1}, ..., y_1, x) = g(y_{i-1}, s_i, c_i)
\end{equation}
where, $y_i$ is the readout, $c_i$ is the context from the encoder and $s_i = f(s_{i-1}, y_{i-1}, c_i)$, is the hidden state of decoder RNN (See \cite{bahdanau2014neural} for details).

However, as discussed in Section \ref{intro}, sequential attention provides the decoder with local context \cite{yang2016encode}. Additionally, providing a semantic input which is closer to language space, as context to the decoder significantly improves on the capability of the model \cite{you2016image}. Our work closely brings these advances together in a Memory Networks framework \cite{weston2014memory,sukhbaatar2015end, kumar2015ask}. While we introduce Key-Value Memory Networks \cite{miller2016key} in a multimodal setting, there are several other key differences from previous works. For one, to our knowledge this is the first work which introduces video captioning in light of Memory Networks. This automatically deals with problems of maintaining long-term dependencies in the input stream by explicitly storing image representations. Meanwhile we also tackle the "vanishing gradient problem" typical with training RNN Encoder-Decoder modules for long input sequences.

Key-Value MemNNs \cite{miller2016key} were originally proposed for QA task in the language domain, providing the last time-step hidden state, as input to the classifier. In this work, we address a more complex problem of video captioning by proposing a novel key-addressing scheme (details in Section \ref{keyadd}) and \textit{(key, value)} setup for exploiting the spatio-temporal structures. The model tracks the attention distribution at previous time steps, thereby providing a strong context on where to attend on the complete video sequence. This implicitly provides a global temporal structure at each readout. While similar in motivation to \cite{you2016image, rocktaschel2015reasoning}, the model architecture and domain of application, especially on capturing global temporal dynamics in videos as opposed to images or entailment, is significantly different.

\section{Key-Value Memory Networks for videos} \label{keyval}
Our work is based on the encoder-decoder framework \cite{cho-al-emnlp14, bahdanau2014neural, DBLP:journals/corr/XuBKCCSZB15, vinyals2015show,karpathy2015deep}, in a Memory Networks \cite{weston2014memory, miller2016key} setting to generate descriptions of videos. The encoder network learns a mapping from the input sequence to a fixed-length vector representation, which is passed to the decoder to generate output sequences. Similar to standard Encoder-Decoder with soft attention mechanism, our model (see Fig. \ref{fig:vidmem}) comprises an encoder module, key-value memories and a decoder module.

\subsection{Encoder}
The encoder network $E$ maps a given input sequence of images in a video $X=\{I_1, ..., I_T$\} of length $T$ to the corresponding sequence of fixed size context representation vectors. As we are dealing with videos (sequence of images), we define two different encoders to achieve the mapping.

\textbf{CNN Encoder}: Given an input image $I_i \in \mathbb{R}^{NxM}$, the CNN encoders learn a mapping $f$ from image $I_i$ to context representations of size $D$ given by $f : \mathbb{R}^{NxM} \to \mathbb{R}^D$. The output of either fully-connected layers\cite{simonyan2014very} or feature maps of convolutional layers\cite{yao2015describing} of standard ConvNet architectures is considered. 

\textbf{RNN Encoder}: The RNN encoder processes the features extracted from CNN Encoder of the frames sequentially, generating hidden states $h^e_i$ at each time step which summarizes the sequence of images seen so far, where
\begin{equation}
h^e_i = g(f(I_i), h^e_{i-1})
\end{equation}
While maintaining temporal dependencies, this allows us to map variable length sequences to fixed length context vectors. In this work, we use modified version of LSTM \cite{hochreiter1997long} unit, as proposed in \cite{zaremba2014recurrent} to implement $g$. The RNN Encoder allows the model to capture the temporal variation between frames and take the ordering of actions and events into consideration. Implicitly, this also helps in preserving high level information about motion in the video \cite{ballas2015delving}. We extract the features from the CNN Encoder, and pass these extracted feature vectors through the RNN Encoder. 

\subsection{Key-Value Memories}
The model is built around a Key-Value Memory Network \cite{miller2016key} with memory slots as key-value pairs $(k_1, v_1), ..., (k_T, v_T)$. The keys and values serve the purpose of transforming visual space context into language space, and effectively capture the relationships between the visual features and textual descriptions. The memory definition, addressing and reading schema is outlined below:

\textbf{Keys $(K)$}: Using CNN Encoder, visual context $k_i$ is generated for each frame $I_i$ of the video. These appearance feature vectors are passed through a RNN Encoder to incorporate sequential structure (video being a sequence of images), and hidden state $h^e_i$ at each timestep is extracted as key $k_i$, 
given by $k_i = h^e_i$.

\textbf{Values $(V)$}: For each image frame $I_i$, a semantic embedding $v_i$ representing the textual meaning of a particular frame/key is generated. It is difficult to jointly learn visual-semantic embedding in Encoder-Decoder models, with supervisory signal only from annotated descriptions \cite{you2016image}. To mitigate this, we explicitly precompute semantic embeddings corresponding to individual frames in the video. In our case, we obtained $v_i$ from a pretrained model $\psi$ which jointly models visual and semantic embedding for images \cite{DBLP:journals/corr/XuBKCCSZB15, johnson2015densecap, vinyals2015show}, given by $v_i = \psi(I_i)$.
Now, for each frame $I_i$ in the video, we have a key-value memory slot $(k_i, v_i)$.

\textbf{Key Addressing}: This corresponds to the soft-attention mechanism deployed to assign a relevance probability $\alpha_i$ to each of the memory slots. These relevance probabilities are used for value reading. We have introduced a new Key Addressing scheme which is described in Section \ref{keyadd}.

\textbf{Value Reading}: The value reading of the memory slots is the weighted sum of the key-value feature vectors: $\phi_t(K)$ and $\phi_t(V)$ at each time step. $\phi_t(K)$ is used for key addressing at the next time step(details in Section \ref{keyadd}) and $\phi_t(V)$ is passed as input to the decoder RNN for generating the next word.
\begin{equation}
	\begin{split}
        \phi_t(K) = \sum^T_{i=1} \alpha_i^{(t)} k_i,  
        \phi_t(V) = \sum^T_{i=1} \alpha_i^{(t)} v_i
	\end{split}
\end{equation}

\subsection{Decoder}
Recurrent Neural Networks is used as decoder because they have been widely used for natural language generation tasks like machine translation, image captioning and video description generation. Since, vanilla RNNs are difficult to train for long range dependencies as they suffer from the \textit{vanishing gradient problem} \cite{bengio1994learning}, Long Short Term Memory(LSTM) \cite{hochreiter1997long} is used. The LSTM units are capable of memorizing context for longer period of time using controllable memory units.

The LSTM model has a memory cell $c_t$ in addition to the hidden state $h_t$ in RNNs, which effectively summarizes the information observed up to that time step. There are primarily three gates which control the flow of information i.e (input, output, forget). The input gate $i_t$ controls the current input $x_t$, forget gate $f_t$ adaptively allows to forget old memory and output gate $o_t$ decides the extent of transfer of cell memory to hidden state. The recurrences at the decoder in our case are defined as:
\begin{align}
    	\mathbf i_t & = \sigma(\mathbf W_i\mathbf h_{t-1} + \mathbf U_i\x_t + \mathbf A_i\mathbf \phi_t(V) + \mathbf b_i) \\
		\mathbf f_t & = \sigma(\mathbf W_f\mathbf h_{t-1} + \mathbf U_f\x_t + \mathbf A_f\mathbf \phi_t(V) + \mathbf b_f) \\
		\mathbf o_t & = \sigma(\mathbf W_o\mathbf h_{t-1} + \mathbf U_o\x_t + \mathbf A_o\mathbf \phi_t(V) + \mathbf b_o) \\
		\mathbf{\tilde{c}_t} & = \tanh(\mathbf W_c\mathbf h_{t-1} + \mathbf U_c\x_t + \mathbf A_c\mathbf \phi_t(V) + \mathbf b_c) \\
		\mathbf c_t & =  \mathbf i_t \odot \mathbf{\tilde{c}_t} + \mathbf f_t \odot \mathbf c_{t-1}\\
        \mathbf h_t & = \mathbf o_t \odot \mathbf c_t
\end{align}
where $\odot$ is an element wise multiplication, $\sigma$ is the sigmoidal non-linearity. $\mathbf W_x, \mathbf U_x, \mathbf A_x$ and $\mathbf b_x$, are the weight matrices for the previous hidden state, input, value context and bias respectively.

Following standard sequence-to-sequence models with generative decoders, we apply a single layer network on the hidden state $h_t$ followed by softmax function to get the probability distribution over the set of possible words.
\begin{align} 
	\mathbf p_t & = \textnormal{softmax}(\mathbf U_p\mathbf [h_t,x_t,\phi_t(V)]  + \mathbf b_p)
\end{align}
Here $p_t$ is the probability distribution over the vocabulary for sampling the current word and [...] denotes vector concatenation. Sentences with high probability are found using Beam Search\cite{sutskever2014sequence}.


\section{Key Addressing} \label{keyadd}

Soft attention mechanism have been successful in image captioning\cite{DBLP:journals/corr/XuBKCCSZB15} and video captioning \cite{yao2015describing} because they focus on the most important segments, and weights them accordingly. Previous work based on soft attention mechanism\cite{yao2015describing} use the decoder's hidden state $h_t$ to find attention weights of each memory unit. We propose a new key addressing mechanism which looks at the previous attention distribution over keys in addition to $h_t$ to select relevant frames for generating the next word. The attention distribution over keys denotes the importance of frames attended so far and the hidden state of the decoder summarizes the previously generated words. This allows us to take into consideration the previously generated words, the attention distribution at previous time steps and the individual key representations $k_i$'s to find relevance score for keys.

\begin{figure}
\centering
\includegraphics[scale=0.5]{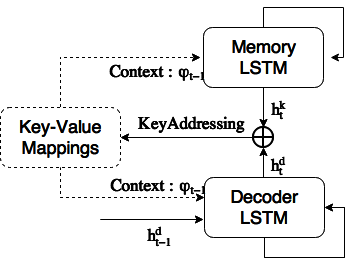}
\caption{The key addressing LSTM is shown here. The memory LSTM updates its hidden state using the last attention distribution over keys. The new hidden state is used with decoder's last hidden state to get new relevance scores which are combined with values and passed to the decoder to generate the next word.}
\label{fig:keyaddlstm}
\end{figure}

We experiment with two different Key-Addressing methods. In first method, we use the previous weighted sum of the keys $\phi_{t - 1}(K)$ directly to find next step attention distribution. In second method, we have a Key-Addressing RNN (referred to as Memory LSTM in Fig. \ref{fig:keyaddlstm}) which takes previous value read over keys $\phi_{t - 1}(K)$ as input.
\begin{equation}
    \mathbf h^k_t = \mathbf f^k(\phi_{t - 1}(K), h^k_{t - 1})
\end{equation}
where $f^k$ is the recurrent unit. For first method, $h^k_t$ is essentially $\phi_{t - 1}(K)$. The next step attention weights $\alpha_i^t$ are obtained using the hidden state $\mathbf h^k_t$ of this RNN-LSTM. The hidden state of Key-Addressing RNN at initial time step is the mean-pooled average of all the keys.

The query vector $q$ is a weighted combination of the decoder and key-addressing hidden states. It summarizes the frames seen so far and the generated outputs. 
\begin{equation}
	q = W_kh^k_t + W_dh_{t - 1}
\end{equation}
For obtaining the attention weights, the relevance score $e^t_i$ of i-th temporal feature $k_i$ is obtained using the decoder RNN hidden state $h_{t - 1}$, key addressing RNN hidden state $h^k_{t}$ and the i-th key vector $k_i$:
\begin{equation}
	e_i^{t} = \mathbf w_{t} \tanh(\mathbf q + \mathbf U_ak_{i})
\end{equation}
where $\mathbf w_{t}$, $\mathbf W_d$, $\mathbf W_k$ and $\mathbf U_a$ are parameters of the model.

These relevance scores are normalised using a softmax function to obtain the new attention distribution $\alpha^t$, where:
\begin{equation}
	\alpha_i^t = exp\{e_i^t\}/\sum^N_{j = 1} exp\{e_j^t\}
\end{equation}


The segregation of the vision and language components into key-value pairs provides a better context for the RNN decoder. Also, the explicit memory structure provides access to the image frames at all time steps allowing the model to assign weights to the key-frames without losing information.

\section{Experimental Setup}
\subsection{Dataset}
\textbf{Youtube2Text} The proposed approach is benchmarked on the Youtube2Text\cite{chen2011collecting} dataset which consists of 1,970 Youtube videos with multiple descriptions annotated through Amazon Mechanical Turk. The videos are generally short (9 seconds on an average), and depict a single activity. Activities depicted are open domain ranging from everyday objects to animals, scenarios, actions, landscapes. etc. The dataset consists of 80,839 annotations with an average of 41 annotations per clip and 8 words per sentence respectively. The training, validation and test sets have 1,200, 100 and 670 videos respectively which is exactly the same splits as in previous work on video captioning \cite{yao2015describing,ballas2015delving,pan2015hierarchical}.
\begin{table*}[t] \label{table}
\caption{Experiment results on the Youtube2Text Dataset. }
\begin{center}
\begin{tabular}{l||c|c|c|c|c}
Model& BLEU@4& METEOR& CIDEr&Feat.&Fine\\
\hline
\hline
VGG-Encoder&0.404& 0.295&0.515&No&No\\
GoogLeNet-Encoder&0.427& 0.303&0.534&No&No\\
t-KeyAddressing &0.436&0.308&0.545&No&No\\
m-KeyAddressing (Memory LSTM) &\textbf{0.457}&\textbf{0.319}&\textbf{0.573}&No&No\\
\hline
Enc-Dec Basic(Yao et al. \cite{yao2015describing})& 0.3869 & 0.2868& 0.4478&No&No\\
GoogLeNet + HRNE(Pan et al. \cite{pan2015hierarchical})&0.438&\textbf{0.321}&.&No&No\\
LSTM-E(VGG + C3D)(Pan et al. \cite{pan2015jointly})&0.453&0.310&.&No&No\\
\hline
 C3D(Yao et al. \cite{yao2015describing})& 0.4192 & 0.2960.& 0.5167&Yes&No\\ 
VGG + C3D + p-RNN(Yu et al.\cite{yu2015video}& \textbf{0.499}&\textbf{0.326}&-&Yes&No\\

S2VT(Venugopalan et al. \cite{venugopalan15iccv})&-&.298&-&Yes&No\\
GRU-RCN(Ballas et al. \cite{ballas2015delving})&\textbf{0.490}&0.3075&\textbf{0.5937}&Yes&Yes\\

\end{tabular}
\end{center}
\end{table*}

\textbf{Key-Value Memories}
We select 28 equally spaced frames and pass them through a pretrained VGG-16\cite{simonyan2014very} and GoogleNet\cite{szegedy2015going} because of their state of the art performance in object detection on Imagenet\cite{imagenet_cvpr09} database. For an input image of size $W X H$, visual features with shape $(\lfloor\frac{W}{16}\rfloor,\lfloor\frac{W}{16}\rfloor,C)$ with $C$ as 512 are extracted from the \emph{conv5\_3} layer of VGG-16.  We simply average over the feature maps which results in a feature vector of size $C$. The visual features extracted from the \emph{pool5/7x7\_s1} layer of GoogLeNet is a 1024 dimensional vector. The feature vectors are either directly used as keys or are passed to encoder RNN to generate keys as described in Section \ref{keyval}.

The values are generated from a pre-trained Densecap \cite{johnson2015densecap} model, which jointly models the task of object localization and textual description. The model identifies salient regions in an image and generates a caption for each of these regions. 
We extract the output of Recognition Network which is encoded as region codes of size $B x D$, where $B$ is the number of salient regions or boxes, and $D$ is the representation with dimension $4096$. Along with the features, a score $S$ is assigned to each of the regions which denotes its confidence. A weighted sum of features of top 5 scores is calculated to get values.

\textbf{Preprocessing}: The video descriptions are tokenized using the wordpunct\_tokenizer from the NLTK toolbox\cite{Loper:2002:NNL:1118108.1118117}. The number of unique words were 15,903 in the Youtube2Text dataset.

\subsection{Model Specifications}
We test on four different variations of the model which help us identify changes in architecture that lead to large improvements on the evaluation metric. \textit{VGG-Encoder} uses features encoded from the last convolution layer in VGG-16 \cite{simonyan2014very} network, and
\textit{GoogLeNet-Encoder} uses features extracted from GoogLeNet\cite{szegedy2015going} as input to the model. There is no Key Addressing in the above two models which means attention weights are obtained using last hidden state of decoder. \textit{t-KeyAddressing} extends the \textit{GoogLeNet-Encoder} by addressing keys using the previous attention distribution over keys. Finally, \textit{m-KeyAddressing} addresses keys using key addressing RNN instead of using the last attention distribution.

\subsection{Model Comparisons}
We compare the model performance with previous state of the art approaches and some strong baselines. Pan et al. \cite{pan2015jointly} explicitly learn a visual-semantic joint embedding model for exploiting the relationship between visual features and generated language, which is then used in a encoder-decoder framework. Yao et al.\cite{yao2015describing} utilizes a temporal attention mechanism for global attention apart from local attention using 3-D Convolution Networks. Ballas et al.\cite{ballas2015delving} proposed an encoder to learn spatial-temporal features across frames, introducing a variant GRU with convolution operations (GRU-RCN). In the current state-of-art Yu et al. \cite{yu2015video} models the decoder as a paragraph generator, describing the videos over multiple sentences using stacked LSTMs.

\subsection{Evaluation Metrics}
We evaluate our approach using standard evaluation metrics to compare the generated sequences with the human annotations, namely BLEU \cite{papineni2002bleu}, METEOR \cite{lavie2014meteor} and CIDEr\cite{vedantam2015cider}. We use the code accompanying the Microsoft COCO Evaluation script \cite{chen2015microsoft} to obtain the results reported in the paper.

\subsection{Training Details}
The model predicts the next output word conditioning on previously generated words and the input video. Thus, the goal is to maximize the log likelihood of the loss function:
\begin{equation}
	L = \frac{1}{N}\sum^N_{i = 1} \sum^{|y^i|}_{j = 1} \textnormal{log\ } p(y^i_j | y^i_{<j}, \mathbf x^n, \textbf{$\theta$})
\end{equation}
where N is the total number of video-description pairs and length of each description $y^i$ is $|y^i|$. Here $x^n$ refers to the input video provided as context to the decoder. We train our network parameters $\theta$ through first order stochastic gradient-based optimization with an adaptive learning rate using the Adadelta \cite{zeiler2012adadelta} optimizer. The batch size is set to be 64 and we optimize hyper-parameters, which include number of hidden units in Decoder LSTM, key addressing LSTM, learning rate and word embedding dimension for the log loss using random search \cite{bergstra2012random}.

\subsection{Results} \label{results}
In the first block of Table \ref{table}, we present the performances of different variations of the model followed by results of prior work in subsequent lines. The \textit{VGG-Encoder} model outperforms S2VT \cite{venugopalan15iccv} and the Basic Enc-Dec model \cite{yao2015describing} on all three metrics, which shows that it is beneficial to use Key-Value Memory Networks in a multimodal setting. We observe that using features from pretrained GoogleNet further improves the results. Using our approach, we further outperform the Enc-Dec model \cite{yao2015describing}.

Results on \textit{t-KeyAddressing} and \textit{m-KeyAddressing} shows further boost in performances on all the metrics demonstrating the effectiveness of using Key Addressing scheme. \textit{m-KeyAddressing} outperforms Pan et al. \cite{pan2015jointly} by a significant margin on BLEU@4. While the improvements on METEOR are significant compared to \textit{t-KeyAddressing}, \cite{pan2015hierarchical} 
performs slightly better. In the current setting, our model is unable to outperform Yu et al.\cite{yu2015video} and Ballas et al.\cite{ballas2015delving}. It must be noted that using more sophisticated regularizers for training the decoder, as proposed in \cite{ba2016layer} and using a better encoder\cite{ballas2015delving} should lead to increased evaluation scores.

In Table \ref{table} we also provide comparison on whether the models use finetuning on the CNN encoder (represented by \textit{Fine}) or if they use external features, like on action recognition, optical flow (represented by \textit{Feat}). It is to be noted, that we do not finetune the encoder
as compared to \cite{ballas2015delving}, which finetunes the encoder CNN on UCF101 action recognition set\cite{soomro2012ucf101}. Also, no additional features are extracted for gaining more information about motion, actions etc. as in \cite{yu2015video}, \cite{ballas2015delving}, \cite{venugopalan15iccv}.

\begin{figure}
\centering
\includegraphics[scale=0.24]{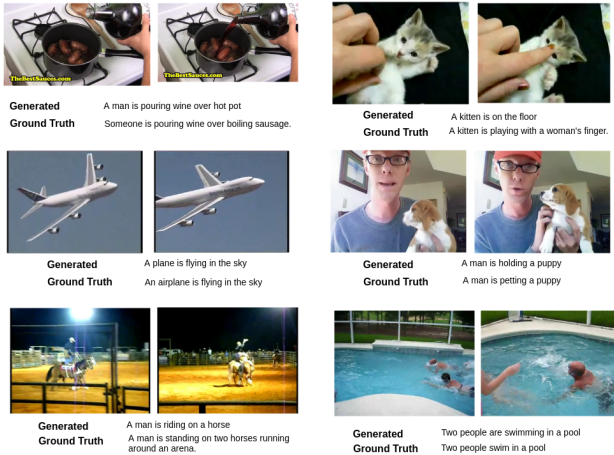}
\caption{Samples generated on the Youtube2Text dataset.}
\label{fig:samples}
\end{figure}

Fig \ref{fig:samples} shows examples of some of the input frames and generated outputs, along with ground truths. Some of the examples demonstrate that the model is able to infer the activities from the video frames, like "swimming", "riding" and "flying" which is distributed across multiple frames. 

\section{Conclusion}
We demonstrate the potential of Memory Networks, specifically Key-Value Memory Networks for video captioning task by decomposing memory into visual and language components as key-value pairs. This paper also proposes a key addressing system for dealing with sequence-to-sequence models, which considers the previous attention distribution over the keys to calculate the new relevance scores. Experiments done on the proposed model outperform strong baselines across several metrics. To the best of our knowledge this is the first proposed work for video-captioning in a Memory Networks setting, and does not rely heavily on annotated videos to generate intermediate semantic-embedding for supporting the decoder. Further work would be exploring the effectiveness of the model on longer videos and generating fine-grained descriptions with more sophisticated decoders.

{\small
\bibliographystyle{ieee}
\bibliography{egbib}
}

\end{document}